\documentclass[runningheads]{llncs}
\usepackage[T1]{fontenc}
\usepackage{graphicx}
\usepackage{color}
\usepackage{textcomp}
\usepackage{booktabs}
\usepackage{multirow}
\usepackage{xcolor}
\usepackage{threeparttable}  
\usepackage{tikz}
\usepackage{amsmath}
\usetikzlibrary{arrows.meta}
\usepackage{stfloats}

\begin{document}
\title{Multiclass Hate Speech Detection with RoBERTa-OTA: Integrating Transformer Attention and Graph Convolutional Networks}
\titlerunning{Multiclass Hate Speech Detection with RoBERTa-OTA}
%
\author{Mahmoud Abusaqer \and Jamil Saquer}
\authorrunning{M. Abusaqer et al.}
%
\institute{Missouri State University, Springfield, MO, 65897, USA \\
\email{\{ma7956s,jamilsaquer\}@missouristate.edu}}
\maketitle              
%


\begin{abstract}
Multiclass hate speech detection across demographic categories remains computationally challenging due to implicit targeting strategies and linguistic variability in social media content. Existing approaches rely solely on learned representations from training data, without explicitly incorporating structured ontological frameworks that can enhance classification through formal domain knowledge integration. We propose RoBERTa-OTA, which introduces ontology-guided attention mechanisms that process textual features alongside structured knowledge representations through enhanced Graph Convolutional Networks. The architecture combines RoBERTa embeddings with scaled attention layers and graph neural networks to integrate contextual language understanding with domain-specific semantic knowledge. Evaluation across 39,747 balanced samples using 5-fold cross-validation demonstrates significant performance gains over baseline RoBERTa implementations and existing state-of-the-art methods. RoBERTa-OTA achieves 96.04\% accuracy compared to 95.02\% for standard RoBERTa, with substantial improvements for challenging categories: gender-based hate speech detection improves by 2.36 percentage points while other hate speech categories improve by 2.38 percentage points. The enhanced architecture maintains computational efficiency with only 0.33\% parameter overhead, providing practical advantages for large-scale content moderation applications requiring fine-grained demographic hate speech classification.

\keywords{Hate speech detection  \and Graph neural networks \and Ontology integration \and Multiclass classification \and Transformer models.}
\end{abstract}
\section{Introduction}

The exponential growth of social media has reshaped digital communication patterns, enabling global connectivity while simultaneously creating new challenges for online safety and content moderation. Balancing free expression with community protection has made automated hate speech detection a critical research area~\cite{gillespie2018custodians}.

Early detection systems relied on lexicon-based and classical machine learning approaches such as Support Vector Machines, Naive Bayes, and ensemble methods~\cite{davidson2017hate,waseem2016hateful}. While effective on curated datasets, these methods struggled with semantic nuance, implicit references, and evolving language patterns. 

The advent of deep learning marked a paradigm shift in natural language processing. CNNs, LSTMs, and attention mechanisms improved performance across social media datasets, and the introduction of transformer architectures such as BERT and RoBERTa set new state-of-the-art results. RoBERTa's optimized training procedures improved robustness over BERT~\cite{liu2019roberta}, with transformer-based models consistently exceeding 90\% accuracy and F1-scores on hate speech detection benchmarks~\cite{abusaqer2025efficient,devlin2019bert}. 

Recent research has explored specialized transformer adaptations for abusive language detection. Yin et al.~\cite{yin2023annobert} combined annotator profiles, label descriptions, and semantic knowledge with Collaborative Topic Regression and BERT to improve detection of difficult posts. 

Mathew et al.~\cite{mathew2021hatexplain} proposed explainable hate speech detection using BERT with attention visualization.
However, most transformer-based approaches continue to address hate speech detection as binary classification problems, limiting their applicability to fine-grained demographic targeting scenarios.

Graph-based methods have been explored to capture structured relationships between hateful concepts. Mishra et al.~\cite{mishra2019abusive} demonstrated that Graph Convolutional Networks effectively model relationships between concepts in abusive language. Wang et al.~\cite{wang2020sosnet} introduced SOSNet, a graph-based framework for fine-grained demographic targeting detection, achieving 92.70\% accuracy and 92.58\% F1-score with SBERT+SOSNet, and 94.38\% accuracy and 94.44\% F1-score using BOW features with XGBoost. Ousidhoum et al.~\cite{ousidhoum2019multilingual} extended this line of work to multilingual, multi-aspect hate speech analysis.

Multiclass detection introduces unique challenges, as different demographic groups exhibit distinct linguistic patterns. While explicit, identity-based hate (e.g., targeting age or religion) is detected reliably, implicit targeting (e.g., gender-based hate) relies on coded language that standard transformers often miss~\cite{abusaquerICMLA25}. These challenges motivate our approach, which integrates structured ontological knowledge with transformer attention to better capture subtle demographic targeting.

Our RoBERTa baseline (95.02\% accuracy, 95.04\% F1-score) and RoBERTa-OTA (96.04\% accuracy, 96.06\% F1-score) outperformed SOSNet's best results on the same dataset, demonstrating the benefit of combining transformer-based contextual learning with structured knowledge for fine-grained hate speech classification.

\section{Literature Review}

Early research relied on hand-crafted features with machine learning classifiers. Davidson et al.~\cite{davidson2017hate} employed TF-IDF features with linear SVM, while Waseem and Hovy~\cite{waseem2016hateful} combined n-gram features with demographic targeting indicators. Burnap and Williams~\cite{burnap2015cyber} added part-of-speech and sentiment features, and Gitari et al.~\cite{gitari2015lexicon} used lexicon-based sentiment dictionaries. These approaches captured some hateful patterns but struggled with semantic and contextual subtleties.

Deep learning models further advanced the field. Badjatiya et al.~\cite{badjatiya2017deep} compared CNN, LSTM, and FastText embeddings, showing consistent improvements over classical approaches. Agrawal and Awekar~\cite{agrawal2018deep} achieved better performance with bidirectional LSTM models and attention mechanisms, solidifying deep learning as the standard for hate speech detection.

Transformer models brought a major leap forward. Devlin et al.~\cite{devlin2019bert} showed that BERT's bidirectional attention achieved state-of-the-art results, later improved by RoBERTa's optimized training~\cite{liu2019roberta}. 
Mathew et al.~\cite{mathew2021hatexplain} demonstrated that transformer attention can also be used for interpretability, offering insights into model decisions alongside strong predictive performance.

Despite these advances, most transformer-based studies remain focused on binary classification. Wang et al.~\cite{wang2020sosnet} addressed this limitation with SOSNet for multiclass detection, achieving strong performance but relying heavily on graph-based representations. Our RoBERTa and RoBERTa-OTA models surpassed these results, highlighting the promise of transformer–ontology integration for fine-grained demographic targeting tasks. Ousidhoum et al.~\cite{ousidhoum2019multilingual} further emphasize the importance of multi-aspect, multilingual analysis for building robust systems.

Overall, current literature shows that while transformers dominate binary hate speech detection, multiclass demographic targeting remains underexplored, presenting an opportunity for more nuanced approaches that combine contextual embeddings with structured knowledge~\cite{abusaqer2025efficient}.

\section{Problem Definition and Research Questions}

We formally define the multiclass hate speech problem as follows: given a dataset $D = \{(x_i, y_i)\}_{i=1}^{n}$ where $x_i$ represents text and $y_i \in \{\text{Age}, \text{Ethnicity}, \text{Gender}, \\ \text{Religion}, \text{Other Hate}\}$, we aim to develop a classifier $f: X \rightarrow Y$ that accurately distinguishes between these categories while maintaining balanced performance across all groups.

We investigate three questions:
\begin{enumerate}
    \item Do ontology-guided transformer architectures outperform standard transformer baselines and state-of-the-art SOSNet results for distinguishing between specific hate speech categories?
    \item Which demographic targeting patterns are most challenging to classify, and how does performance vary across hate speech categories?
    \item What computational overhead do enhanced architectures introduce, and do performance gains justify the additional resource requirements?
\end{enumerate}

\section{Dataset and Preprocessing}

Our experimental evaluation utilizes the fine-grained hate speech dataset derived from the SOSNet framework introduced by Wang et al.~\cite{wang2020sosnet}, which provides comprehensive coverage of demographic-targeted hate speech patterns across social media platforms. The dataset comprises 39,747 samples systematically extracted from a larger collection of 47,692 hate speech tweets, with careful filtering applied to ensure balanced representation across five distinct hate speech categories. Following the SOSNet methodology, this filtering process removes the generic ``none\_hate'' category to focus specifically on distinguishing between different forms of hate speech, creating a classification task that better reflects real-world content moderation challenges. This also allows us to compare our results with SOSNet's. Table~\ref{tab:dataset_distribution} shows class distributions of the dataset.

\begin{table}[htbp]
\centering
\caption{Dataset Statistics: 5-Class Hate Speech Distribution}
\label{tab:dataset_distribution}
\begin{tabular}{lccc}
\toprule
\textbf{Hate Speech Class} & \textbf{Samples} & \textbf{Percentage (\%)}  \\
\midrule
Religion & 7,998 & 20.1\\
Age & 7,992 & 20.1 \\
Gender & 7,973 & 20.1  \\
Ethnicity & 7,961 & 20.0 \\
Other Hate & 7,823 & 19.7  \\
\midrule
\textbf{Total} & \textbf{39,747} & \textbf{100.0}  \\
\bottomrule
\end{tabular}
\end{table}

Each hate speech category exhibits distinct linguistic characteristics reflecting different targeting strategies. Age-based hate speech employs generational stereotypes and age-related mockery, while gender-based harassment includes appearance-focused attacks and sexualized content. Ethnicity-based hate speech utilizes cultural stereotypes and racial slurs requiring specialized cultural understanding. Religion-based targeting involves theological criticism and sectarian conflicts with specific religious terminology. The ``Other Hate'' category captures diverse forms targeting socioeconomic status, physical disabilities, sexual orientation, and additional characteristics beyond conventional demographic boundaries.

\subsection{Linguistic Validation and Class Characterization}

To validate the necessity of fine-grained classification approaches, we conducted comprehensive linguistic analysis across all demographic categories. This analysis demonstrates that distinct hate speech categories exhibit linguistic differences that justify sophisticated classification architectures, providing quantitative evidence that RoBERTa-OTA's performance improvements reflect genuine model capability rather than dataset artifacts.

Our linguistic analysis employs part-of-speech tagging using spaCy's \\ \texttt{en\_core\_web\_sm} model to extract structural features across all 39,747 samples. The feature extraction pipeline computes character count, token count, noun frequency, verb frequency, and URL detection using regular expression patterns, capturing both surface-level characteristics and deeper syntactic patterns that distinguish different forms of demographic targeting.

The linguistic analysis reveals substantial heterogeneity across hate speech categories, as summarized in Table~\ref{tab:linguistic_features}. Religion-based hate speech demonstrates the highest linguistic complexity, averaging 198.0 characters with 10.1 nouns and 39.6 tokens, indicating extensive theological terminology usage. Age-based targeting exhibits considerable narrative content with 173.5 characters and 37.0 tokens per sample. Other forms of hate speech display the most heterogeneous patterns, averaging only 85.7 characters but showing the highest URL usage (0.18 per sample), suggesting diverse propagation strategies including external content sharing.

\begin{table}[htbp]
\centering
\caption{Average Linguistic Features Across Hate Speech Categories}
\label{tab:linguistic_features}
\begin{tabular}{lrrrrr}
\toprule
\textbf{Class} & \textbf{URLs} & \textbf{Chars} & \textbf{Tokens} & \textbf{Nouns} & \textbf{Verbs} \\
\midrule
Religion & 0.07 & 198.0 & 39.6 & 10.1 & 4.6 \\
Age & 0.03 & 173.5 & 37.0 & 7.8 & 4.8 \\
Ethnicity & 0.04 & 139.3 & 30.4 & 7.8 & 3.4 \\
Gender & 0.11 & 136.4 & 29.3 & 7.4 & 3.2 \\
Other Hate & 0.18 & 85.7 & 18.4 & 4.2 & 2.1 \\
\bottomrule
\end{tabular}
\end{table}

\begin{figure}[htbp]
\centering
\includegraphics[scale=0.513]{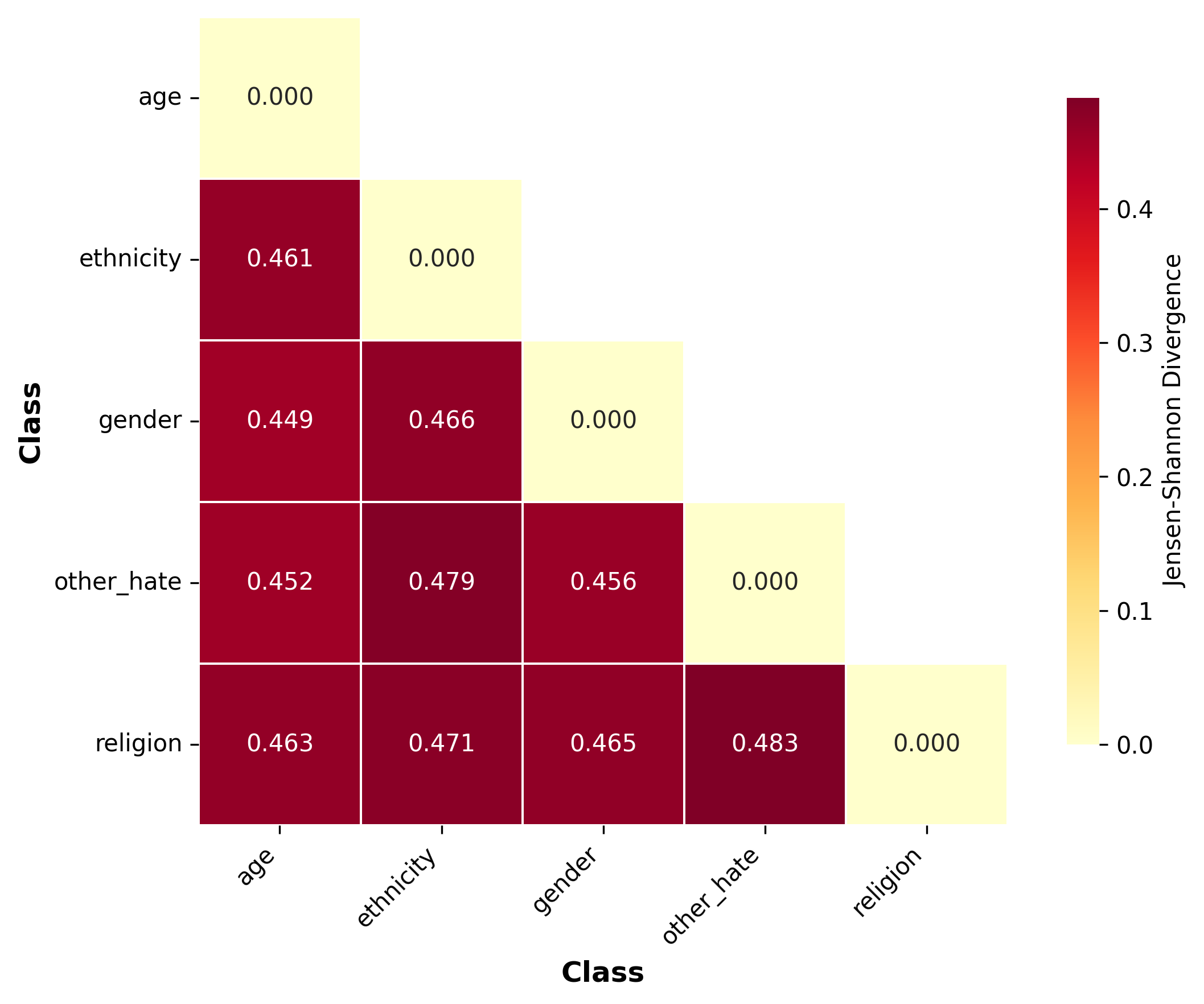}
\caption{Pairwise Jensen-Shannon divergence between hate speech classes.}
\label{fig:divergence_analysis}
\end{figure}

Figure~\ref{fig:divergence_analysis} presents pairwise Jensen-Shannon divergence scores computed from unigram language models for each hate speech category, where values range from 0 (identical distributions) to 1 (maximally different distributions). The analysis reveals substantial lexical separation between all class pairs, with divergence scores ranging from 0.449 to 0.483 (mean = 0.466), demonstrating robust linguistic separation necessary for fine-grained classification approaches. Other hate and religion-based hate speech exhibit maximum divergence (0.483), while age-based and gender-based harassment show minimum divergence (0.449), yet this value still represents substantial lexical distinction supporting the necessity of sophisticated attention mechanisms.

\subsection{Preprocessing Pipeline}

We implement minimal preprocessing to retain contextual information crucial for distinguishing between different hate speech types, as aggressive preprocessing can eliminate subtle linguistic cues that differentiate demographic targeting patterns. Text normalization includes fundamental cleaning procedures such as removing excessive whitespace, handling special characters, and standardizing Unicode encoding to ensure consistent text representation.

The preprocessing workflow follows established protocols adapted from recent hate speech detection research. Structural cleaning operations include user mention removal for privacy, and HTML tag removal to clean web scraping artifacts. Content normalization procedures involve converting text to lowercase, expanding contractions using standardized libraries, and converting emojis to textual representations to preserve emotional context while enabling consistent processing.

Our preprocessing strategy deliberately preserves emoticons, punctuation patterns, and certain capitalization structures that may convey important sentiment or emotional intensity information relevant to hate speech classification. This approach recognizes that aggressive text normalization can eliminate crucial linguistic signals that distinguish between different forms of demographic targeting, particularly in social media contexts where unconventional language use often carries semantic significance.

\section{Methodology}

Our experimental methodology compares two architectures for 5-class hate speech detection: a RoBERTa baseline and our novel RoBERTa-OTA enhancement. Both implementations utilize PyTorch with identical optimization parameters and undergo evaluation through 5-fold cross-validation with stratified sampling to ensure fair comparison across fine-grained demographic hate speech categories.

\subsection{Baseline Architecture: RoBERTa}

RoBERTa represents an optimized variant of BERT with enhanced pretraining procedures, including dynamic masking and improved training strategies. Our implementation employs RoBERTa-base with 124.6M parameters as the baseline architecture for 5-class hate speech detection. We utilize the pretrained model with an enhanced classification head featuring multiple linear layers and dropout regularization, mapping the 768-dimensional contextual representations to five demographic hate speech categories through progressive dimensionality reduction.

\subsection{Novel Architecture: RoBERTa-OTA}

Our proposed RoBERTa-OTA (RoBERTa with Ontology-guided Transformer Attention) represents an advanced architecture that combines transformer-based text processing with domain-specific knowledge integration through graph neural networks. This dual-stream approach builds upon RoBERTa's enhanced language understanding capabilities while incorporating structured ontological knowledge about hate speech categories and their semantic relationships, enabling more nuanced classification across fine-grained demographic targeting patterns.

The RoBERTa-OTA architecture incorporates three critical design innovations. First,  \textbf{enhanced 3-layer Graph Convolutional Network (GCN)} processes structured demographic relationships through systematic progression (6→64→64→32 dimensions) using GCNConv layers with ReLU activation and layer normalization. Second, the \textbf{feature integration mechanism} concatenates RoBERTa's 768-dimensional textual representations with 32-dimensional ontological features, creating 800-dimensional combined representations for classification. Third, \textbf{deep classifier network} employs three linear transformations (800→400→200→5) with batch normalization, layer normalization, and progressive dropout rates (0.3, 0.2, 0.1) specifically optimized for fine-grained demographic classification.

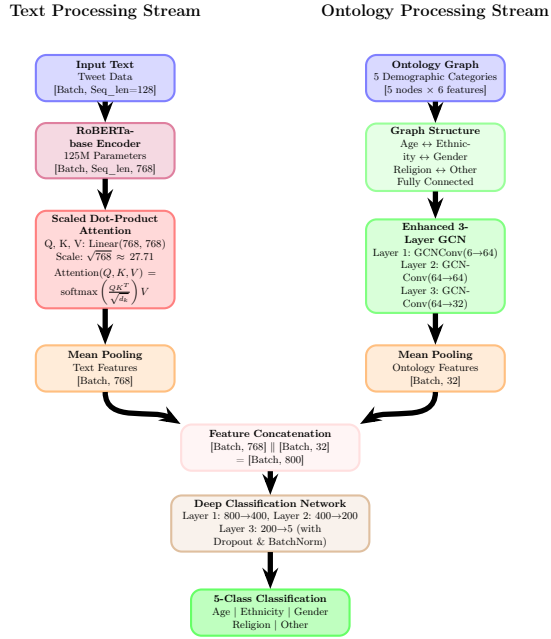
\begin{figure}[htbp]
\centering
\begin{tikzpicture}[scale=0.439, transform shape]

\node at (-5, 15) {\Large\textbf{Text Processing Stream}};
\node at (5, 15) {\Large\textbf{Ontology Processing Stream}};

\node[rectangle, rounded corners=4pt, minimum width=4.2cm, minimum height=1.4cm,
      fill=blue!15, draw=blue!50, thick, text width=3.8cm, align=center] 
      (input-text) at (-5, 13) {
    \textbf{Input Text}\\
    Tweet Data\\
    {[Batch, Seq\_len=128]}
};

\node[rectangle, rounded corners=4pt, minimum width=4.2cm, minimum height=1.4cm,
      fill=purple!15, draw=purple!50, thick, text width=3.8cm, align=center] 
      (roberta) at (-5, 10.8) {
    \textbf{RoBERTa-base Encoder}\\
    125M Parameters\\
    {[Batch, Seq\_len, 768]}
};

\node[rectangle, rounded corners=4pt, minimum width=4.2cm, minimum height=2.0cm,
      fill=red!15, draw=red!50, thick, text width=3.8cm, align=center] 
      (attention) at (-5, 7.5) {
    \textbf{Scaled Dot-Product}\\
    \textbf{Attention}\\
    Q, K, V: Linear(768, 768)\\
    Scale: $\sqrt{768} \approx 27.71$\\[0.1cm]
    $\text{Attention}(Q,K,V) =$\\
    $\text{softmax}\left(\frac{QK^T}{\sqrt{d_k}}\right)V$
};

\node[rectangle, rounded corners=4pt, minimum width=4.2cm, minimum height=1.4cm,
      fill=orange!15, draw=orange!50, thick, text width=3.8cm, align=center] 
      (text-pool) at (-5, 4.2) {
    \textbf{Mean Pooling}\\
    Text Features\\
    {[Batch, 768]}
};

\node[rectangle, rounded corners=4pt, minimum width=4.2cm, minimum height=1.4cm,
      fill=blue!15, draw=blue!50, thick, text width=3.8cm, align=center] 
      (input-ontology) at (5, 13) {
    \textbf{Ontology Graph}\\
    5 Demographic Categories\\
    {[5 nodes × 6 features]}
};

\node[rectangle, rounded corners=4pt, minimum width=4.2cm, minimum height=1.4cm,
      fill=green!10, draw=green!40, thick, text width=3.8cm, align=center] 
      (graph-vis) at (5, 10.6) {
    \textbf{Graph Structure}\\
    Age {$\leftrightarrow$} Ethnicity {$\leftrightarrow$} Gender\\
    Religion {$\leftrightarrow$} Other\\
    Fully Connected
};

\node[rectangle, rounded corners=4pt, minimum width=4.2cm, minimum height=1.7cm,
      fill=green!15, draw=green!50, thick, text width=3.8cm, align=center] 
      (gnn) at (5, 7.3) {
    \textbf{Enhanced 3-Layer GCN}\\
    Layer 1: GCNConv(6→64)\\
    Layer 2: GCNConv(64→64)\\
    Layer 3: GCNConv(64→32)
};

\node[rectangle, rounded corners=4pt, minimum width=4.2cm, minimum height=1.4cm,
      fill=orange!15, draw=orange!50, thick, text width=3.8cm, align=center] 
      (ontology-pool) at (5, 4.2) {
    \textbf{Mean Pooling}\\
    Ontology Features\\
    {[Batch, 32]}
};

\node[rectangle, rounded corners=4pt, minimum width=5.4cm, minimum height=1.4cm,
      fill=pink!15, draw=pink!50, thick, text width=5.0cm, align=center] 
      (concatenation) at (0, 1.8) {
    \textbf{Feature Concatenation}\\
    {[Batch, 768]} $\parallel$ {[Batch, 32]}\\
    = {[Batch, 800]}
};

\node[rectangle, rounded corners=4pt, minimum width=6.0cm, minimum height=1.7cm,
      fill=brown!10, draw=brown!50, thick, text width=5.6cm, align=center] 
      (classifier) at (0, -0.5) {
    \textbf{Deep Classification Network}\\
    Layer 1: 800→400, Layer 2: 400→200\\
    Layer 3: 200→5 (with Dropout \& BatchNorm)
};

\node[rectangle, rounded corners=4pt, minimum width=4.8cm, minimum height=1.4cm,
      fill=green!25, draw=green!60, thick, text width=4.4cm, align=center] 
      (output) at (0, -3.2) {
    \textbf{5-Class Classification}\\
    Age | Ethnicity | Gender\\
    Religion | Other
};

\draw[-{Stealth[length=8pt,width=6pt]}, very thick, line width=2pt] (input-text) -- (roberta);
\draw[-{Stealth[length=8pt,width=6pt]}, very thick, line width=2pt] (roberta) -- (attention);
\draw[-{Stealth[length=8pt,width=6pt]}, very thick, line width=2pt] (attention) -- (text-pool);

\draw[-{Stealth[length=8pt,width=6pt]}, very thick, line width=2pt] (input-ontology) -- (graph-vis);
\draw[-{Stealth[length=8pt,width=6pt]}, very thick, line width=2pt] (graph-vis) -- (gnn);
\draw[-{Stealth[length=8pt,width=6pt]}, very thick, line width=2pt] (gnn) -- (ontology-pool);

\draw[-{Stealth[length=8pt,width=6pt]}, very thick, line width=2pt] (text-pool) to[out=270,in=150] (concatenation.north west);
\draw[-{Stealth[length=8pt,width=6pt]}, very thick, line width=2pt] (ontology-pool) to[out=270,in=30] (concatenation.north east);

\draw[-{Stealth[length=8pt,width=6pt]}, very thick, line width=2pt] (concatenation) -- (classifier);
\draw[-{Stealth[length=8pt,width=6pt]}, very thick, line width=2pt] (classifier) -- (output);

\end{tikzpicture}
\caption{RoBERTa-OTA Architecture: Dual-Stream Processing Framework for Multiclass Hate-Speech Detection with Ontology-Guided Attention}
\label{fig:roberta-ota-architecture}
\end{figure}

The RoBERTa-OTA architecture implements a sophisticated dual-stream processing framework that handles both textual features and structured knowledge representations in parallel as depicted in Figure~\ref{fig:roberta-ota-architecture}.

\subsubsection{Text Processing Stream}

The text processing stream utilizes pre-trained RoBERTa-base for robust contextual embeddings, enhanced through additional scaled dot-product attention layers specifically optimized for hate speech detection patterns. The scaled dot-product attention mechanism computes refined attention scores using the standard formulation:

\begin{equation}
\text{Attention}(Q, K, V) = \text{softmax}\left(\frac{QK^T}{\sqrt{d_k}}\right)V
\end{equation}

where $Q$, $K$, and $V$ represent learned linear projections of the input embeddings, and $d_k$ denotes the dimension of the keys corresponding to RoBERTa's hidden size of 768. This mechanism enables focused attention on hate speech-relevant features while maintaining the contextual understanding capabilities of the base transformer architecture.

\subsubsection{Ontology Processing Stream}

Our hate speech ontology represents domain-specific knowledge through a structured graph comprising five nodes corresponding to the target demographic categories (age, ethnicity, gender, religion, \\other\_hate). Each node is characterized by a six-dimensional feature vector encoding conceptual attributes derived from linguistic analysis and established hate speech research that capture the semantic properties and targeting characteristics of each category.

The node feature vectors encode the following semantic dimensions: demographic targeting (binary), cultural identity targeting (binary), gender-related characteristics (binary), religious/belief targeting (binary), linguistic complexity intensity (continuous), and targeting diversity (continuous). These dimensions reflect both the targeting mechanisms and the observed linguistic patterns identified in our comprehensive dataset analysis.

Our ontological framework follows systematic design principles based on comprehensive analysis of our 39,747-sample dataset. The six-dimensional feature vectors encode conceptual attributes derived from empirical linguistic analysis and established hate speech research that capture the semantic properties and targeting characteristics of each category. Each dimension represents empirically-grounded characteristics: demographic targeting (binary indicator of direct population targeting), cultural identity targeting (binary indicator reflecting cultural/ethnic elements), gender-related characteristics (binary indicator for gender-specific patterns), religious/belief targeting (binary indicator for theological elements), linguistic complexity (continuous measure reflecting observed character-per-sample statistics), and targeting diversity (continuous measure of attack vector breadth within categories). 

For instance, the religion node features [1.0, 1.0, 0.0, 1.0, 0.9, 0.8] reflect empirical observations: direct demographic and cultural targeting, religious characteristics, highest linguistic complexity (0.9, corresponding to 198.0 characters per sample with extensive theological terminology), and substantial targeting diversity (0.8) across sectarian boundaries. The fully-connected graph structure enables systematic relationship modeling between categories while maintaining interpretable encodings aligned with computational linguistics research. While cross-linguistic evaluation remains future work, the framework's grounding in demographic targeting mechanisms and linguistic complexity patterns provides a systematic foundation for adaptation to diverse cultural contexts.

\begin{table}[htbp]
\centering
\caption{Ontological Node Feature Vectors for Hate Speech Categories}
\label{tab:ontology_features}
\begin{tabular}{lcccccc}
\toprule
\textbf{Category} & \textbf{Demo.} & \textbf{Cult.} & \textbf{Gend.} & \textbf{Relig.} & \textbf{Complex.} & \textbf{Divers.} \\
\midrule
Age       & $[1.0$ & 0.0 & 0.0 & 0.0 & 0.7 & $0.3]$ \\
Ethnicity & $[1.0$ & 1.0 & 0.0 & 0.0 & 0.6 & $0.7]$ \\
Gender    & $[1.0$ & 0.0 & 1.0 & 0.0 & 0.6 & $0.5]$ \\
Religion  & $[1.0$ & 1.0 & 0.0 & 1.0 & 0.9 & $0.8]$ \\
Other   & $[0.5$ & 0.5 & 0.5 & 0.5 & 0.4 & $0.9]$ \\
\bottomrule
\end{tabular}
\begin{tablenotes}
\footnotesize
\item Demo.\;=\;Demographic targeting; Cult.\;=\;Cultural identity; Gend.\;=\;Gender-related; Relig.\;=\;Religious/belief; Complex.\;=\;Linguistic complexity; Divers.\;=\;Targeting diversity
\end{tablenotes}
\end{table}

Table~\ref{tab:ontology_features} summarizes the complete ontological feature representation for all hate speech categories. The demographic dimension identifies primary population targeting (1.0 for direct demographic attacks), while cultural and religious dimensions capture identity-based targeting mechanisms. Linguistic complexity values reflect the observed analytical patterns from our dataset, with religion exhibiting the highest complexity (0.9) corresponding to extensive theological terminology, and other categories showing the lowest complexity (0.4) reflecting diverse but simpler targeting patterns. Targeting diversity measures the breadth of attack vectors within each category, with other forms of hate speech demonstrating the highest diversity (0.9) due to their heterogeneous nature spanning multiple personal characteristics beyond conventional demographic boundaries.

The graph connectivity follows a fully-connected structure where each category node maintains bidirectional edges with all other nodes, enabling the capture of inter-category relationships and semantic overlaps between different forms of demographic targeting. This structured representation enables the Graph Convolutional Network (GCN) to learn nuanced relationships between hate speech categories while maintaining interpretable feature encodings that align with established hate speech research and observed linguistic patterns in our comprehensive dataset analysis.

The ontological knowledge stream employs a three-layer GCN that processes structured relationships between different hate speech concepts, demographic categories, and their linguistic manifestations. Each GCN layer implements the standard graph convolution operation:
\begin{equation}
X^{(l+1)} = \tilde{D}^{-\frac{1}{2}}\tilde{A}\tilde{D}^{-\frac{1}{2}}X^{(l)}W^{(l)}
\end{equation}
where $\tilde{A} = A + I$ represents the adjacency matrix with added self-connections, $\tilde{D}$ denotes the degree matrix, $X^{(l)}$ represents node features at layer $l$, and $W^{(l)}$ constitutes the learnable weight matrix. The GCN architecture employs a first layer with input dimension 6 and hidden dimension 64, followed by a second layer maintaining hidden dimension 64, and a third layer with output dimension 32. The first two layers apply ReLU activation, layer normalization, and dropout (0.1) for training stability, while the final layer produces raw outputs for subsequent classification processing.

\subsubsection{Feature Integration and Classification}

Feature integration occurs through a carefully designed concatenation mechanism that combines mean-pooled representations from both processing streams. The text stream contributes contextual embeddings enhanced through scaled attention, while the ontological stream provides structured domain knowledge encoded through graph convolutional operations. The integration mechanism follows:

\begin{equation}
F = \text{mean}(\text{attention\_output}) \parallel \text{mean}(\text{GCN\_features})
\end{equation}

where $\parallel$ denotes concatenation and mean pooling operations ensure fixed-size representations suitable for subsequent classification processing. The integrated features undergo dropout regularization with rate 0.3 before passing through the final linear classification layer that maps combined textual and ontological patterns to the five hate speech categories.

\subsection{Training Configuration and Optimization}

All models are implemented using PyTorch on two NVIDIA Ampere A100 GPUs (6912 CUDA cores, 40GB RAM each). We utilize the Transformers library for pre-trained model access and PyTorch Geometric for graph neural network components. Text processing employs RoBERTa's tokenizer with a maximum sequence length of 128 tokens, applying necessary padding and truncation with consistent handling across all architectures.

Both architectures employ AdamW optimizer with identical parameters: learning rate 1e-5, beta values (0.9, 0.999), and weight decay 1e-5. Training proceeds for a maximum of 20 epochs with batch size 16, implementing early stopping with patience of 3 epochs monitoring validation F1-score. Both architectures use cross-entropy loss with label smoothing ($\alpha=0.1$) for enhanced regularization.

We use 5-fold stratified cross-validation to maintain consistent class distribution. Accuracy and F1-weighted scores are the primary evaluation metrics, measuring overall correctness and balanced performance across demographic groups. Supplementary metrics include precision, and recall. Training efficiency evaluation incorporates per-fold training times and GPU memory utilization.

\section{Results and Discussion}

\begin{table}[!t]
\centering
\caption{Comprehensive Performance Analysis Across 5-Class Hate Speech Detection}
\label{tab:per_class_analysis}
\begin{tabular*}{\columnwidth}{@{\extracolsep{\fill}}lcccc|cccc}
\toprule
\multirow{2}{*}{\textbf{Class}} & \multicolumn{4}{c}{\textbf{RoBERTa Baseline}} & \multicolumn{4}{c}{\textbf{RoBERTa-OTA}} \\
\cmidrule(lr){2-5} \cmidrule(lr){6-9}
 & \textbf{F1} & \textbf{Acc} & \textbf{Prec} & \textbf{Rec} & \textbf{F1} & \textbf{Acc} & \textbf{Prec} & \textbf{Rec} \\
\midrule
Age & 98.63 & 98.42 & 98.85 & 98.42 & 98.75 & 98.84 & 99.08 & 98.42 \\
Ethnicity & 98.41 & 98.10 & 98.72 & 98.10 & 98.44 & 97.84 & 99.05 & 97.84 \\
Gender & 90.70 & 91.50 & 90.03 & 91.50 & 93.06 & 93.43 & 92.71 & 93.43 \\
Religion & 98.28 & 97.52 & 99.05 & 97.52 & 98.62 & 98.56 & 98.68 & 98.56 \\
Other Hate & 88.94 & 89.33 & 88.68 & 89.33 & 91.32 & 91.87 & 90.82 & 91.87 \\
\midrule
\textbf{Weighted Avg} & \textbf{95.04} & \textbf{95.02} & \textbf{95.12} & \textbf{95.02} & \textbf{96.06} & \textbf{96.04} & \textbf{96.09} & \textbf{96.04} \\
\midrule
\textbf{SOSNet~\cite{wang2020sosnet}} & \multicolumn{4}{c}{\textbf{94.44 F1, 94.38 Acc}} & \multicolumn{4}{c}{\textbf{Previous State-of-the-Art}} \\
\bottomrule
\end{tabular*}
\end{table}

\begin{table}[!t]
\centering
\caption{Computational Efficiency Comparison}
\label{tab:efficiency_comparison}
\begin{tabular*}{\columnwidth}{@{\extracolsep{\fill}}lccccc}
\toprule
\textbf{Model} & \textbf{FLOPs} & \textbf{Params} & \textbf{GPU Mem} & \textbf{Size} & \textbf{Time} \\
 & \textbf{(Billions)} & \textbf{(Millions)} & \textbf{(GB)} & \textbf{(MB)} & \textbf{(min/fold)} \\
\midrule
RoBERTa Baseline & 31.9 & 124.65 & 2.6 & 475.5 & 24.5 \\
RoBERTa-OTA & 35.2 & 125.06 & 3.1 & 477.1 & 27.5 \\
\midrule
\textbf{Overhead} & \textbf{+10.3\%} & \textbf{+0.33\%} & \textbf{+19.2\%} & \textbf{+0.34\%} & \textbf{+12.2\%} \\
\bottomrule
\end{tabular*}
\end{table}

Our experimental evaluation demonstrates that RoBERTa-OTA consistently outperforms the RoBERTa baseline across all evaluation metrics. Table~\ref{tab:per_class_analysis} presents the comprehensive performance analysis, while Table~\ref{tab:efficiency_comparison} examines computational costs. We discuss these results in relation to our three research questions.

RoBERTa-OTA achieves 96.04\% accuracy compared to the RoBERTa baseline's 95.02\%, representing a 1.02 percentage point improvement. The F1-weighted score increases from 95.04\% to 96.06\%, while precision improves from 95.12\% to 96.09\%. We notice similar improvement across all hate speech categories. The consistency across metrics indicates genuine performance gains rather than optimization artifacts.

Our results establish new state-of-the-art performance for fine-grained hate speech detection. The previous best approach, SOSNet using BOW+XGBoost, achieved 94.38\% accuracy and 94.44\% F1-score on the same dataset. RoBERTa-OTA's 96.04\% accuracy and 96.06\% F1-score represent 1.66 and 1.62 percentage points improvement over existing methods, demonstrating the effectiveness of combining transformer attention with structured ontological knowledge.

Table~\ref{tab:per_class_analysis} reveals distinct performance patterns across hate speech categories. Religion, age, and ethnicity-based hate speech achieve high F1-scores (98.62\%, 98.75\%, and 98.44\% respectively), reflecting their explicit targeting language and clear discriminative signals. Gender-based and other hate speech present greater classification challenges but benefit most from ontology-guided attention. Gender-based hate speech improves from 90.70\% to 93.06\% F1-score (+2.36 percentage points), while other hate speech increases from 88.94\% to 91.32\% (+2.38 percentage points). These substantial improvements occur precisely where implicit targeting strategies and coded language create ambiguity for standard transformer approaches.

Table~\ref{tab:efficiency_comparison} demonstrates that RoBERTa-OTA introduces modest architectural overhead. Parameter count increases minimally from 124.65M to 125.06M (0.33\% overhead), while peak GPU memory usage increases from 2.6GB to 3.1GB. Model size increases by only 1.6MB, maintaining deployment feasibility across standard hardware configurations. Training time shows modest computational overhead, with average fold time increasing from 24.5 to 27.5 minutes (12.2\% increase). However, RoBERTa-OTA demonstrates superior convergence efficiency, requiring fewer total epochs (29 vs 31 epochs across 5-fold cross-validation) due to the ontology-guided learning process. The FLOPs increase of 10.3\% reflects additional processing for ontological integration. The performance improvements of 1.02 percentage points in accuracy combined with the convergence efficiency make RoBERTa-OTA attractive for applications requiring accurate fine-grained hate speech detection.

While the overall accuracy improvement appears incremental (+1.02 percentage points), this analysis misrepresents the practical value proposition of our approach. The computational overhead yields disproportionate improvements precisely where content moderation systems struggle most: gender-based hate speech detection improves by 2.36 percentage points (90.70→93.06\% F1) and other hate speech by 2.38 percentage points (88.94→91.32\% F1). These categories represent the most challenging detection scenarios in real-world deployment, where implicit targeting strategies deliberately evade standard detection methods. In production environments processing millions of messages daily, a 2.36\% improvement in gender-based hate speech detection translates to thousands of correctly identified harmful posts that would otherwise remain undetected.
Furthermore, our architectural efficiency analysis reveals favorable trade-offs beyond simple FLOPs comparison. While computational operations increase by 10.3\%, parameter overhead remains minimal at only 0.33\% (125.06M vs 124.65M parameters), maintaining deployment feasibility on standard hardware. Critically, RoBERTa-OTA demonstrates superior convergence efficiency, requiring fewer training epochs (29 vs 31 across 5-fold cross-validation) due to ontology-guided learning, effectively offsetting training time overhead through faster convergence. The 19.2\% memory increase (3.1GB vs 2.6GB) remains within practical deployment constraints, while model size increases by only 1.6MB. For applications prioritizing accuracy on challenging hate speech categories over raw computational efficiency, these targeted performance gains justify the modest computational overhead.

\subsection{Robustness Under Social Media Text Perturbations}

\begin{table*}[!t]
\centering
\caption{Model Robustness Under Social Media Text Perturbations}
\label{tab:robustness}
\begin{tabular}{@{}l@{\hspace{2.5em}}l@{\hspace{1.5em}}cc@{\hspace{1.5em}}cc@{\hspace{1.5em}}cc@{}}
\toprule
\multirow{2}{*}{\textbf{Perturbation}} & \multirow{2}{*}{\textbf{Level}} & \multicolumn{2}{c@{\hspace{1.5em}}}{\textbf{RoBERTa Baseline}} & \multicolumn{2}{c@{\hspace{1.5em}}}{\textbf{RoBERTa-OTA}} & \multicolumn{2}{c}{\textbf{Improvement}} \\
\cmidrule(lr){3-4} \cmidrule(lr){5-6} \cmidrule(lr){7-8}
\textbf{Type} & & \textbf{Acc (\%)} & \textbf{F1 (\%)} & \textbf{Acc (\%)} & \textbf{F1 (\%)} & \textbf{$\Delta$ Acc} & \textbf{$\Delta$ F1} \\
\midrule
Clean Text & -- & 97.50 & 97.49 & 98.53 & 98.53 & +1.03 & +1.04 \\
\midrule
\multirow{3}{*}{Character} & 5\% & 92.53 & 92.56 & 94.13 & 94.19 & +1.60 & +1.63 \\
\multirow{3}{*}{Deletion} & 10\% & 85.50 & 85.45 & 88.47 & 88.67 & +2.97 & +3.22 \\
 & 15\% & 76.73 & 76.17 & 79.23 & 79.64 & +2.50 & +3.47 \\
\addlinespace
\multirow{3}{*}{Character} & 5\% & 92.23 & 92.25 & 93.83 & 93.92 & +1.60 & +1.67 \\
\multirow{3}{*}{Substitution} & 10\% & 86.27 & 86.25 & 87.47 & 87.84 & +1.20 & +1.59 \\
 & 15\% & 79.33 & 79.02 & 80.97 & 81.77 & +1.64 & +2.75 \\
\addlinespace
\multirow{3}{*}{Character} & 5\% & 93.37 & 93.37 & 95.37 & 95.40 & +2.00 & +2.03 \\
\multirow{3}{*}{Insertion} & 10\% & 87.93 & 87.93 & 90.73 & 90.90 & +2.80 & +2.97 \\
 & 15\% & 81.30 & 81.20 & 85.47 & 85.82 & +4.17 & +4.62 \\
\addlinespace
\multirow{3}{*}{Word} & 10\% & 94.20 & 94.22 & 94.37 & 94.45 & +0.17 & +0.23 \\
\multirow{3}{*}{Deletion} & 20\% & 91.20 & 91.29 & 92.33 & 92.50 & +1.13 & +1.21 \\
 & 30\% & 86.97 & 87.10 & 87.43 & 87.76 & +0.46 & +0.66 \\
\addlinespace
\multirow{3}{*}{Abbreviation} & 10\% & 97.10 & 97.08 & 98.37 & 98.36 & +1.27 & +1.28 \\
\multirow{3}{*}{Injection} & 20\% & 96.87 & 96.84 & 98.37 & 98.36 & +1.50 & +1.52 \\
 & 30\% & 96.87 & 96.84 & 98.37 & 98.36 & +1.50 & +1.52 \\
\addlinespace
\multirow{3}{*}{Slang} & 10\% & 97.53 & 97.52 & 98.53 & 98.53 & +1.00 & +1.01 \\
\multirow{3}{*}{Substitution} & 20\% & 97.53 & 97.52 & 98.53 & 98.53 & +1.00 & +1.01 \\
 & 30\% & 97.53 & 97.52 & 98.53 & 98.53 & +1.00 & +1.01 \\
\bottomrule
\end{tabular}
\end{table*}

Social media text contains misspellings, abbreviations, and slang that challenge hate speech detection systems. We evaluated model robustness using six perturbation types on 3,000 sampled tweets: character deletion, substitution, and insertion (5\%, 10\%, 15\%), word deletion (10\%, 20\%, 30\%), abbreviation injection (10\%, 20\%, 30\%), and slang substitution (10\%, 20\%, 30\%).

Table~\ref{tab:robustness} shows RoBERTa-OTA consistently outperforms the baseline across all perturbations. Character-level noise causes the largest degradation for both models. At 15\% character deletion, RoBERTa F1-score drops to 76.17\% (21.32 point decrease), while RoBERTa-OTA maintains 79.64\% (18.89 point decrease), demonstrating a 3.47 percentage point F1 advantage. Character insertion reveals the most substantial robustness gap: RoBERTa-OTA achieves 85.82\% F1 versus the baseline's 81.20\% at 15\% perturbation, yielding a 4.62 percentage point F1 improvement. This enhanced resilience stems from ontological knowledge integration, which provides semantic grounding when surface forms become corrupted.

Both models exhibit remarkable stability under social media linguistic variations. At 30\% abbreviation substitution, RoBERTa-OTA maintains 98.36\% F1-score (only 0.17 point drop from clean text), while the baseline achieves 96.84\% F1 (0.65 point drop). Slang substitution shows even greater resilience: both architectures maintain clean text F1 performance levels with minimal degradation across all intensity levels. The consistent F1-score advantages across perturbation types demonstrate that RoBERTa-OTA's robustness benefits extend uniformly across all five hate speech categories, validating its effectiveness for practical multiclass deployment in noisy social media environments.

\section{Conclusion and Future Work}

Our comprehensive evaluation of RoBERTa-OTA demonstrates significant advances in multiclass hate speech detection through ontology-guided attention mechanisms. The dual-stream architecture achieved 96.04\% accuracy and 96.06\% weighted F1-score, outperforming the RoBERTa baseline by 1.02 percentage points in both metrics and surpassing the previous state-of-the-art SOSNet approach by 1.62 (F1) and 1.66 (accuracy) percentage points. The integration of scaled attention with enhanced Graph Convolutional Networks proved particularly effective for challenging categories, improving gender-based hate speech detection by 2.36 percentage points and other hate speech by 2.38 percentage points, while adding only 0.33\% parameter overhead.

For future work, we plan to evaluate RoBERTa-OTA on multilingual datasets to assess its effectiveness across diverse cultural and linguistic contexts. We also aim to explore efficiency optimizations to reduce computational overhead without compromising accuracy. The strong performance of our ontology-guided approach motivates further investigation into advanced knowledge integration techniques, particularly for capturing evolving patterns in online hate speech.

%
%
%
\bibliographystyle{splncs04}
\bibliography{mybibliography}
\end{document}